\documentclass{article}

\usepackage{microtype}
\usepackage{graphicx}
\usepackage{subfigure}
\usepackage{booktabs} 

\usepackage{hyperref}

\usepackage[accepted]{icml2024}

\usepackage{amsmath}
\usepackage{amssymb}
\usepackage{mathtools}
\usepackage{amsthm}

\usepackage{multirow}

\usepackage[capitalize,noabbrev]{cleveref}

\theoremstyle{plain}

\theoremstyle{definition}

\theoremstyle{remark}

\usepackage[textsize=tiny]{todonotes}

\icmltitlerunning{
Submission and Formatting Instructions for ICML 2024
}

\begin{document}

\twocolumn[
\icmltitle{
Information Leakage from Embedding in Large Language Models
}



\icmlsetsymbol{equal}{*}

\begin{icmlauthorlist}
\icmlauthor{Zhipeng Wan}{equal,comp}
\icmlauthor{Anda Cheng}{equal,comp}
\icmlauthor{Yinggui Wang}{comp}
\icmlauthor{Lei Wang}{comp}
\end{icmlauthorlist}

\icmlaffiliation{comp}{Ant Group}

\icmlcorrespondingauthor{Yinggui Wang}{wyinggui@gmail.com}

\icmlkeywords{Machine Learning, ICML}

\vskip 0.3in
]



\printAffiliationsAndNotice{\icmlEqualContribution} 

\begin{abstract}

The widespread adoption of large language models (LLMs) has raised concerns regarding data privacy. This study aims to investigate the potential for privacy invasion through input reconstruction attacks, in which a malicious model provider could potentially recover user inputs from embeddings.
We first propose two base methods to reconstruct original texts from a model's hidden states. 
We find that these two methods are effective in attacking the embeddings from shallow layers, but their effectiveness decreases when attacking embeddings from deeper layers.
To address this issue, we then present Embed Parrot, a Transformer-based method, to reconstruct input from embeddings in deep layers.
Our analysis reveals that Embed Parrot effectively reconstructs original inputs from the hidden states of ChatGLM-6B and Llama2-7B, showcasing stable performance across various token lengths and data distributions. To mitigate the risk of privacy breaches, we introduce a defense mechanism to deter exploitation of the embedding reconstruction process. Our findings emphasize the importance of safeguarding user privacy in distributed learning systems and contribute valuable insights to enhance the security protocols within such environments. 
\end{abstract}

\section{Introduction}
\label{sec:intro}
Despite their exceptional performance across a range of natural language tasks, large language models (LLMs) still necessitate fine-tuning to improve their responsiveness to specific tasks, as highlighted by the work of \cite{chen2023federated}. 
With the popularity of LLMs, some researchers have started to explore fine-tuning LLMs based on Federated Learning(FL) which stands as a widely embraced framework for decentralized machine learning model training\cite{zhang2023building,zhang2023federated,babakniya2023slora}. FL's core objective is to achieve highly accurate model training without compromising the privacy of client data. However, recent works \cite{zhao2020idlg,zhu2019deep,phong2017privacy} have shown that attackers can recover the client data by applying a reconstruction attack on the gradient updates sent from the client during training.
In other cases, 
some systems utilize LLMs to store auxiliary data in densely embedded vector databases\cite{borgeaud2022improving, yao2023react}, and users of these systems inject knowledge into the LLMs by inserting retrieved documents into the language model's hints. In these cases, the data owner only sends an embedding of the text data to the third-party service, but not the text itself\cite{kiros2015skipthought,le2014distributed}. However, the privacy threats involved have not been adequately explored. 

The above scenarios involve a privacy issue: 
Can a third-party service recreate the original text via text embedding?
Generally, consider the following setting: If the model provider splits the model structure into two parts, the former part is used by the user to process his input and output the hidden embeddings to the model provider, and then the model provider uses the latter part of the model to process the hidden embeddings and output the final result to the user. Can a malicious model provider recover the user's input based on the hidden embeddings passed by the user and violate the user's privacy?

In this work, we explore the intrinsic privacy concerns associated with the use of large language models (LLMs), particularly within the contexts of federated learning (FL) and embedded vector databases. We investigate the potential for third-party services or malicious model providers to reconstruct original text from text embeddings, posing a significant privacy risk to users. Our exploration is motivated by the emergent risks as LLMs increasingly handle sensitive data within distributed and collaborative learning environments.
We present Embed Parrot, an innovative approach developed to execute input reconstruction attacks on systems by exploiting their latent representations. Embed Parrot employs a Transformer-based architecture to master the intricate mappings between input data and its corresponding hidden states, thus probing the vulnerability of such systems to disclose sensitive information through obscured model representations.

Our contributions are multifaceted. Firstly, we propose and conduct a comparative analysis of several embedding inversion techniques, including Base Embed Inversion, Finetuned Embed Inversion, Hotmap Embed Inversion, and Embed Parrot. The study evaluates their effectiveness in reconstructing the original inputs of ChatGLM-6B and Llama2-7B. We utilize a robust set of metrics to ensure an exhaustive assessment of the reconstruction quality.
Secondly, we explore the relationship between tokens length and reconstruction performance. Our experiments reveal that Embed Parrot performs consistently across varying tokens lengths, demonstrating its stability as a reconstruction method. The capability of Embed Parrot is tested across different datasets, indicating that its performance is impacted by data distribution, yet it is not confined to a single distribution type.
Finally, we propose simple yet effective defense mechanisms to prevent the misuse of the Embed Parrot. This defense approach offer practical means to safeguard against malicious exploitation of embedding reconstruction capabilities.
Through rigorous experimentation and methodical analysis, our work contributes to the ongoing discourse on securing LLMs against emerging privacy threats in federated learning contexts.

\section{Related Work}
\label{sec:related}

\subsection{Hidden State in the Transformer}
Following the rise of LLM, numerous studies are focused on the interpretability of hidden states of a Transformer forward pass. For instance, the work of \cite{vanaken2020visbert} found that hidden states can reflect the "thought" process of transformer language models by projecting them to the vocabulary space.
Previous work \cite{geva2021transformer} shows that the learned patterns are human-interpretable and that lower layers tend to capture shallow patterns, while upper layers learn more semantic ones. It also shows how the fully connected blocks of transformer LMs add information to the model's residual stream and eventually make it to the final predictions.
These works suggest hidden states can be viewed as the "language" of transformer language models, not just numbers or vectors. 

\subsection{Information Leakage In LLMs}
Nowadays, models are usually pre-trained on massive data collected from various sources, including data collected by institutions, data retrieved by web crawlers and so on\cite{naveed2023comprehensive}. Although people will intentionally avoid personal private data being used for training, the problem of model privacy leakage is still hard to avoid.
For embedding inversion attacks, there are a lot of related studies. \cite{9152761} found the text embeddings from general-purpose language models would capture much sensitive information from the plain text. 
\cite{song2020information} systematically studied information that embeddings might leak. Recently, \cite{Gu2023TowardsSL} proposed a method to recover the target sequences word by word directly by exploiting generative decoders, and \cite{morris2023text} proposed Vec2Text to refine the inverted text sequences iteratively. Similar to this work, \cite{li-etal-2023-sentence} also proposed a method to reconstruct the input sequence based on sentence embeddings.
Gradient leakage problems\cite{NEURIPS2019_60a6c400} are also widely discussed and still rather unexplored for natural language processing. \cite{fowl2023decepticons} exploited characteristics of both the Transformer architecture and the token embedding, separately extracting tokens and positional embeddings to retrieve high-fidelity text. \cite{balunovic2022lamp} proposed a method to reconstruct original text from gradients. While we were writing this paper, we found a new work,\cite{morris2023language} showing a method for recovering unknown prompts given only the model's current distribution output.

\section{Preliminaries}
In this section, we describe relevant background of language models.
\subsection{Language Modeling} \label{PRE:LM}
 Given a sequence of $n$ tokens
$x = \{x_1, x_2, ..., x_n\}$, the language modeling task is to estimate the probability distribution of $P(x)$:
\begin{align}
\log P_\theta(x) &= \sum_{i=1}^{n} \log P_\theta(x_{i}|x_1,...,x_{i-1})
\end{align}
Contemporary LLMs are characterized by their reliance on the Transformer architecture \cite{vaswani2017attention}, which is composed of a highly extensive parameter space, typically encompassing millions or even billions of parameters, represented as $\theta$. 
A language model initially transforms the input token $x$ into a sequence of vectors through the application of a word embedding matrix $W$. 
This matrix resides in the space $R^{|V| \times d}$, where $|V|$ denotes the cardinality of the vocabulary set $V$, and $d$ represents the dimensionality of the embedding, commonly referred to as the hidden dimension. 
The model then computes hidden states, represented as $h_i$, which are conditioned on the preceding tokens $x_1,..., x_{i-1}$. 
These hidden states $h_i$ refer to the final layer's output at the $i$-th position in the sequence, capturing the contextual information up to that point. Using these hidden states, the model predicts the probability of the next token as:
\begin{align}
{{P_\theta(x_{i}|x_1,...,x_{i-1})}=\frac{\exp({h_i^T \cdot W_{x_i}})}{\sum_{j \in V}{\exp({h_i^T \cdot W_{j}})}}}
\end{align}

\subsection{Generation of Intermediate Representations}

The Transformer architecture employs multiple layers to refine input data representations, creating intermediate hidden states. Each layer $i$ processes the preceding layer's outputs, $H^{i-1}$, beginning with initial embeddings ($H^0$) derived from input tokens using an embedding matrix $W$.
At each layer, the hidden states undergo transformation through a multi-head self-attention layer (MHA) , enabling contextual information integration across the sequence. The attention output is then subjected to a position-wise feed-forward neural network (FFN) for further refinement. These processes can be mathematically represented as:
\begin{align}
H^i = \operatorname{FFN}(\operatorname{MHA}(H^{i-1}; \theta_i); \theta_i)
\end{align}

Here, $\theta_i$ includes all parameters of the $i$-th layer, encompassing both the attention and feed-forward networks. These parameters are optimized during training to capture complex patterns in the data. The cascading effect of these layers leads to increasingly abstract representations, essential for the Transformer model's performance in language processing tasks.

\subsection{Token Decoding Function in Language Models}
To elucidate the mechanism by which a tokenizer converts the latent representations generated by a Transformer model into a textual format, it is instructive to delineate a sequence of abstract functions that encapsulate this transformation.
Commence by introducing a function $logits$ that maps the final hidden states to a logits vector. This function embodies the operations of the language modeling head:
\begin{align}
\operatorname{logits}(i) &= H^i \cdot w + b
\end{align}
Here, $w$ denotes the weights of the language modeling head and $b$ represents the bias term. Subsequently, a softmax operation is applied to the logits to derive a probability distribution over the tokens, followed by the selection of the index corresponding to the most probable token:
\begin{align}
\operatorname{input\_ids}(i) &= \arg\max(\operatorname{Softmax}(\operatorname{logits}(i)))
\end{align}
Next, we introduce the decoding function $Decode$, which translates the series of input IDs into coherent human-readable text:
\begin{align}\label{PRE:Output_i}
\operatorname{Output}(i) = \operatorname{Decode}(\operatorname{input\_ids}(i))
\end{align}
By synthesizing the aforementioned functions, the comprehensive procedure for transmuting the hidden states from the final layer (i.e., the N-th layer) into a sequence of human-readable text can be articulated as follows:
\begin{align}
\operatorname{Output}(N) = \operatorname{Decode}(\arg\max(\operatorname{Softmax}(\operatorname{logits}(H^N))))
\end{align}

\section{Method}



\label{sec:method}
We aim to reconstruct the original input text from the hidden states of a specified Transformer layer.
In this section, we introduce three embedding inversion methods to achieve this goal, which are Base Embed Inversion (BEI), Hotmap Embed Inversion (HEI), and Embed Parrot (EP).


\subsection{Base Embed Inversion} \label{method-BEI}


LLMs based on the Transformer architecture generally forward the hidden states produced by the last layer of the Transformer layers to the language modeling head. 
This is done to obtain the probability distribution of predicted tokens, as seen in models like BERT, LLAMA, and others. Similar to \cite{morris2023language}, Base Embed Attack (BEI) expects to use the probability distribution to recover the original text.
Thus, the attack procedure of BEI can be described as metioned in section \ref{PRE:Output_i}:
\begin{align}
    \operatorname{BEI}(H^i) 
    = \operatorname{Decode}(\arg\max(\operatorname{Softmax}(\operatorname{logits}(H^i))))
\end{align}

Here, $H^i$ refers to the hidden states output by the i-th layer of the Transformer structure.
Based on our experimental observations, directly passing the hidden states of any layer to the language modeling heads can to some extent recover the original input.

\subsection{Hotmap Embed Inversion}
Embeddings are functions that map raw input data to low-dimensional vector representations, while preserving important semantic information about the inputs\cite{song2020information}. 
HEI's approach is based on the model's reliance on word embeddings.
Specifically, HEI method is predicated on the maximization of cosine similarity within the embedding space of a pre-trained language model. It focuses on efficiently determining the vocabulary word whose embedding vector is most similar to the hidden states($H^i$) output by the $i$-th layer. To achieve this, the HEI method calculates the pairwise cosine similarities between the target embedding vector and all embedding vectors in the model's vocabulary, while the original embedding weights remain unchanged. The embedding space is represented by the weight matrix $W$ mentioned in \ref{PRE:LM}.
The HEI process can be mathematically formulated as follows:
\begin{align}
\operatorname{HEI}(H^i) = \operatorname{Decode}(\arg\max\left(\frac{W {H^i}^T}{|W| |{H^i}^T|}\right))
\end{align}

\subsection{Embed Parrot}
We empirically find that for BEI and HEI, it is much more difficult to reconstruct the original input from the hidden states of deeper layers than the initial hidden states (as shown in Table \ref{BEI-FEI:llama}).
To solve this issue, we propose an innovative method Embed Parrot which contains a specially trained lightweight model using the Transformer architecture, which is trained on datasets to reconstruct the hidden states of a given layer to the original hidden state. 
Finally, we combined the restored hidden states with HEI's method or BEI's method to obtain an output that is very similar to the semantics of the original input.

Formally, given the original input is $x$, a certain layer $i$ is selected from the transformer layers of the attacked model, and the output embedding of $x$ through layer $i$ of the transformer layers of the attacked model is $H^i_x$, we train Embed Parrot $M$ to generate output embedding $H^i_y$ which is similar with $H^0_x$. $H^0_x$ is also equivalent to the output of the original input $x$ after passing through the word embedding layer of the attacked model. The training objective can be succinctly captured as 
$\min_M |H^i_y - H^0_x|$.

Embed Parrot's main module consists of three parts, which are \texttt{Input-Adapter},
\texttt{Decode Model}, and \texttt{Output-Adapter}. \texttt{Input-Adapter} is a linear layer used to handle problems where the hidden size of $H^i_y$ does not match the hidden size required in the transformer layers of the \texttt{Decode Model} so that the \texttt{Decode Model} can handle the passed embeddings. \texttt{Output-Adapter} is a linear layer used to convert the output of the \texttt{Decode Model} into the hidden size of $H^i_y$. The \texttt{Decode Model} is an only-decoder architecture model, and we choose GPT-2 XL.GPT-2 XL is the 1.5B parameter version of GPT-2 \cite{radford2019language}, a transformer-based language model created and released by OpenAI. The model is a pre-trained model on the English language using a causal language modeling (CLM) objective. We also experimented with other pre-trained models with a similar number of parameters, such as the encoder-only model and encoder-decoder model, but in general, the decoder-only architecture performed best.

We used two datasets in the training process, one for training Embed Parrot and the other to validate the performance of Embed Parrot's reduced inputs. We use the cosine similarity of $H^i_y$ and $H^0_x$ to optimize the loss function of Embed Parrot. 
To enhance the fluency of the outputs generated by the language models, we appended to the loss function the Perplexity (PPL) stemming from the reconstructed sentences when used as inputs to LLMs.

\section{Experiments}
\label{sec:exp}

\subsection{Experimental Setups}

\subsubsection{Datasets}

In this study, we utilized two FinGPT\cite{yang2023fingpt} training datasets: \textit{fingpt-sentiment} and \textit{fingpt-fiqa\_qa}, along with the training split of the \textit{wikitext-2-raw-v1} subset from the Wikitext\cite{merity2016pointer} dataset. The \textit{fingpt-sentiment} is a training dataset designed for sentiment analysis within the financial domain, comprising $76.8K$ lines of data. The \textit{fingpt-fiqa\_qa} dataset is tailored for training question answering models in the financial sector and contains $17.1K$ lines of data. Meanwhile, the \textit{wikitext-2-raw-v1} represents curated content from Wikipedia, consisting of $44.8K$ lines of data. We randomized these datasets and allocated $20\%$ of the data to form separate test sets for each. Predominantly, the training splits derived from \textit{fingpt-sentiment} and \textit{wikitext-2-raw-v1} were employed for fine-tuning or training our models, and the performance of various text restoration methods was assessed across the three derived test sets.

\begin{table*}[!tbp]
\centering
\caption{Comparison of three reconstruction methods. 
For ChatGLM-6B, the methods are applied to reconstruct results from the hidden states output by the 28th layer, whereas for Llama2-7B, the 30th layer's hidden states are utilized.
}
\begin{tabular}{llcccccccc}
\toprule
Model & Metrics & BEI   & HEI & EP  \\
\midrule
\multirow[c]{4}{*}{ChatGLM-6B} & F1 Score & 39.19   & 40.23 & 75.08 \\
 & ROUGE-1 & 42.86   & 41.37 & 75.02  \\
 & ROUGE-L & 39.42   & 38.72 & 69.62 \\
 & Sentence-BERT  & 53.46   & 64.29 & 76.10 \\
\midrule
\multirow[c]{4}{*}{Llama2-7B} & F1 Score & 20.41   & 26.08 & 65.99\\
 & ROUGE-1 & 16.66   & 26.02 &  65.97 \\
 & ROUGE-L & 15.69   & 25.95 &  62.48\\
 & Sentence-BERT  & 22.30   & 50.48 &  68.85 \\
\bottomrule
\end{tabular}\label{all-methods-compare}
\end{table*}

\subsubsection{Models}
Our experiments are performed on two different target models based on the transformer architecture, more specifically, the models we used are Llama2-7B and ChatGLM-6B.
Llama 2 \cite{touvron2023llama}is an auto-regressive language model that uses an optimized transformer architecture. It outperforms other open-source language models on many external benchmarks, including reasoning, coding, proficiency, and knowledge tests.
Llama2-7B is a 7B fine-tuned version, optimized for dialogue use cases and converted for the Hugging Face Transformers format. It has 32 layers with the hidden dimension of 4096, and 32 attention heads.
GLM\cite{du2022glm} is a General Language Model pretrained with an autoregressive blank-filling objective and can be finetuned on various natural language understanding and generation tasks. ChatGLM-6B is an open bilingual language model based on the General Language Model (GLM) framework, with 6.2 billion parameters. It has 28 layers with the same hidden dimension as Llama2-7B, and 32 attention heads.

\subsubsection{Metrics}
We measure the success of our methods based on a comprehensive set of metrics, including the ROUGE family of metrics \cite{lin-2004-rouge}. Specifically, we report the ROUGE-1 and ROUGE-L scores, which assess the overlap of recovered sentences with the original sentences in terms of unigrams and longest common subsequences, respectively. Additionally, we incorporate the F1 Score adopted from the SQuAD benchmark \cite{rajpurkar2016squad}, providing a harmonic mean of precision and recall in our evaluations. Furthermore, we utilize the Sentence-BERT Score \cite{reimers2019sentencebert}, which serves as an additional metric for semantic similarity, capturing the contextual relationship between the compared sentences. These metrics together provide a robust framework for evaluating the performance of our natural language processing methods.

\subsection{Experimental Results}

\begin{table*}[!tbp]
\centering
\caption{Performance of BEI and HEI on Llama2-7B.
The numbers in the first row indicate the layer indexes. 
}
\label{BEI-FEI:llama}
\begin{tabular}{llcccccccc}
\toprule
Method & Metrics & 0 & 5 & 10 & 15 & 20 & 25 & 30 & last \\
\midrule
\multirow[c]{4}{*}{BEI} & F1 Score & 0.17 & 1.98 & 2.46 & 3.25 & 13.10 & 18.10 & 20.41 & 23.01 \\
 & ROUGE-1 & 0.33 & 1.71 & 1.75 & 3.20 & 9.33 & 13.34 & 16.66 & 18.23 \\
 & ROUGE-L & 0.33 & 1.68 & 1.69 & 3.04 & 8.57 & 12.19 & 15.69 & 17.59 \\
 & Sentence-BERT  & 3.00 & 2.31 & 2.99 & 6.62 & 22.61 & 30.95 & 22.30 & 32.41 \\
\midrule
\multirow[c]{4}{*}{HEI} & F1 Score & 85.07 & 73.44 & 65.39 & 61.73 & 53.50 & 39.30 & 26.08 & 16.67 \\
 & ROUGE-1 & 65.22 & 58.93 & 53.09 & 50.11 & 44.38 & 34.39 & 26.02 & 19.51 \\
 & ROUGE-L & 65.23 & 58.95 & 53.07 & 50.04 & 44.40 & 34.37 & 25.95 & 19.13 \\
 & Sentence-BERT  & 90.15 & 84.19 & 80.27 & 77.75 & 72.65 & 61.01 & 50.48 & 25.11 \\
\bottomrule
\end{tabular}
\end{table*}

\begin{table*}
\centering
\caption{Performance of BEI and HEI on ChatGLM-6B at various layers. The numbers in the first row indicate the layer indexes.
}
\begin{tabular}{llcccccccc}
\toprule
 Method & Metrics  & 0 & 5 & 10 & 15 & 20 & 25 & last \\
\midrule
\multirow[c]{4}{*}{BEI} & F1 Score & 70.75 & 68.56 & 67.31 & 74.10 & 65.38 & 49.49 & 39.19 \\
 & ROUGE-1 & 100.00 & 97.41 & 93.26 & 86.62 & 72.33 & 55.46 & 42.86 \\
 & ROUGE-L & 100.00 & 97.47 & 93.29 & 86.64 & 72.41 & 53.16 & 39.42 \\
 & Sentence-BERT  & 71.64 & 70.61 & 70.48 & 77.92 & 71.51 & 64.41 & 53.46 \\
\midrule
 \multirow[c]{4}{*}{HEI} & F1 Score & 99.94 & 96.46 & 93.94 & 91.23 & 60.68 & 51.78 & 40.23 \\
 & ROUGE-1 & 100.00 & 96.72 & 92.17 & 89.89 & 61.05 & 55.28 & 41.37 \\
 & ROUGE-L & 100.00 & 96.76 & 92.18 & 89.87 & 60.98 & 53.69 & 38.72 \\
 & Sentence-BERT  & 99.99 & 98.15 & 97.01 & 95.07 & 67.03 & 71.28 & 64.29 \\
\bottomrule
\end{tabular}\label{FEI-BEI:chatglm}
\end{table*}

\begin{table*}
\centering
\caption{Effectiveness of EP on different datasets and models. 
CP and LP indicate the attack model are trained by incorporating the PPL provided by ChatGLM-6B and Llama2-7B into the loss function as optimization terms, respectively.   
}
\label{EmbedParrot}



\begin{tabular}{lllcccccc}
\toprule
 Testing Dataset& Model & Method  &ROUGE-1 & ROUGE-L & F1 Score & Sentence-BERT  \\
\midrule
\multirow[c]{4}{*}{fingpt-fiqa\_qa} & ChatGLM-6B & EP & 75.08 & 75.02 & 69.62 & 76.10 \\
 & Llama2-7B & EP & 65.99 & 65.97 & 62.48 & 68.85 \\
 & Llama2-7B & EP + CP  & 65.16 & 65.12 & 61.61 & 68.23 \\
 & Llama2-7B & EP + LP & 64.98 & 64.96 & 61.42 & 67.70 \\
\midrule

\multirow[c]{4}{*}{fingpt-sentiment} & ChatGLM-6B & EP & 76.96 & 76.92 & 68.84 & 76.65 \\
 & Llama2-7B &  EP & 67.30 & 67.23 & 61.69 & 71.47 \\
 & Llama2-7B & EP + CP  & 66.57 & 66.51 & 60.91 & 70.74 \\
 & Llama2-7B & EP+LP  & 66.28 & 66.21 & 60.73 & 70.91 \\
\midrule

\multirow[c]{4}{*}{wikitext} & ChatGLM-6B & EP  & 40.08 & 39.98 & 67.24 & 65.45 \\
 & Llama2-7B & EP & 66.03 & 65.84 & 63.55 & 60.46 \\
 & Llama2-7B & EP+CP  & 65.56 & 65.34 & 63.07 & 60.45 \\
 & Llama2-7B & EP+LP  & 65.45 & 65.24 & 62.93 & 60.26 \\
\bottomrule
\end{tabular}

\end{table*}

\begin{table*}
    \centering
    \caption{
    Performance of EP to reconstruct tokens with different length ranges on fingpt-sentiment dataset.
    }
    \label{length-fingpt-sentiment}
    \begin{tabular}{llccccccccc}
    \toprule
     Model&  & Metrics& (0, 8) & [8, 16) & [16, 32) & [32, 64) & [64, 128) & [128, 256) & [256, 512)  \\
    \midrule
    \multirow[c]{4}{*}{ChatGLM-6B}&\multirow[c]{4}{*} & ROUGE-1 & 60.72 & 71.05 & 76.93 & 78.91 & 82.41 & 79.85 & 71.76\\
     & &ROUGE-L & 60.72 & 71.04 & 76.89 & 78.87 & 82.36 & 79.85 & 71.76  \\
     & &F1 Score & 49.26 & 63.12 & 69.03 & 70.40 & 75.43 & 70.43 & 61.75  \\
     & &Sentence-BERT  & 60.13 & 71.75 & 76.69 & 78.06 & 82.48 & 83.17 & 78.87  \\
    \midrule
    \multirow[c]{4}{*}{Llama2-7B(CP)}&\multirow[c]{4}{*} & ROUGE-1 & 68.09 & 68.12 & 66.01 & 66.54 & 67.92 & 63.89 & 53.40  \\
     & &ROUGE-L & 68.09 & 68.10 & 65.97 & 66.45 & 67.73 & 63.46 & 53.40  \\
     & &F1 Score & 67.10 & 65.23 & 60.98 & 59.32 & 62.34 & 58.39 & 52.27  \\
     & &Sentence-BERT  & 73.39 & 70.87 & 70.06 & 70.74 & 74.92 & 74.20 & 76.46  \\
    \midrule
    \multirow[c]{4}{*}{Llama2-7B(LP)}&\multirow[c]{4}{*} & ROUGE-1 & 67.71 & 67.80 & 65.98 & 66.10 & 66.90 & 63.32 & 52.34  \\
     & &ROUGE-L & 67.71 & 67.77 & 65.93 & 66.01 & 66.68 & 62.75 & 52.34  \\
     & &F1 Score & 66.72 & 64.88 & 61.00 & 59.10 & 61.42 & 58.03 & 51.10  \\
     & &Sentence-BERT  & 74.47 & 70.97 & 70.37 & 70.83 & 74.66 & 74.00 & 76.64  \\
    \midrule
    \multirow[c]{4}{*}{Llama2-7B}&\multirow[c]{4}{*} & ROUGE-1 & 67.23 & 68.57 & 66.78 & 67.33 & 68.59 & 64.81 & 53.25  \\
     & &ROUGE-L & 67.23 & 68.55 & 66.74 & 67.25 & 68.39 & 64.24 & 53.25  \\
     & &F1 Score & 66.49 & 65.64 & 61.80 & 60.18 & 63.07 & 59.32 & 51.40  \\
     & &Sentence-BERT  & 74.28 & 71.29 & 70.91 & 71.52 & 75.07 & 74.86 & 79.04  \\
    \bottomrule
    \end{tabular}
\end{table*}

\subsubsection{Main Results} 
The experimental findings, as detailed in Table \ref{FEI-BEI:chatglm} and Table \ref{BEI-FEI:llama}, reveal that the Base Embed Inversion (BEI) method is adept at reconstructing text from the early layers of ChatGLM-6B, but its performance wanes as it traverses into the deeper layers of the model. This trend is in stark contrast with Llama2-7B, where BEI's performance for text reconstruction steadily improves with each subsequent layer. Nonetheless, the best results achieved with Llama2-7B are still inferior to the lowest performance metrics recorded for ChatGLM-6B. The divergent trends underscore the potential impact of the distinct architectural designs of the models on the effectiveness of BEI.

When the Hotmap Embed Inversion (HEI) is applied to ChatGLM-6B, there is a marked enhancement in all metrics for text reconstruction from the hidden states of the initial layers. Despite these gains, HEI shows slight performance deterioration at specific layers, such as the 20th, when contrasted with the BEI approach. In the case of Llama2-7B, HEI demonstrates a clear advantage, delivering superior performance over BEI in nearly all evaluated layers, solidifying its suitability for this particular model. These results highlight the necessity of tailoring inversion strategies to suit the idiosyncrasies of different LLM architectures. We also consider the impact of fine-tuning the models on the efficacy of the BEI and HEI methods, with detailed results presented in the appendix \ref{impacts-of-FT}.

Both BEI and HEI exhibit diminished effectiveness in the deeper layers of the models, a deficiency for which Embed Parrot is specifically designed to address.

We train Embed Parrot for ChatGLM-6B layer 28 and Llama2-7B layer 30 on FinGPT's fingpt-sentiment dataset, respectively. As shown in Table \ref{all-methods-compare}, for Llama2, the highest rouge score of the text restored by the previous methods on the hidden states of layer 30 is only about 0.26, but the highest rouge score of the text restored by the previous methods on the hidden states output by Embed Parrot is 0.659. For ChatGLM-6B, the text score of hidden states restored on the 28th layer increases from 0.54 to 0.75, which fully demonstrates the versatility and effectiveness of Embed Parrot.Table \ref{EmbedParrot} presents the performance of Embed Parrot across varying experimental configurations. It is discernible that Embed Parrot performs optimally on the dataset used for its training, fingpt-sentiment. Subsequently, its performance on the fingpt-fiqa\_qa dataset, which has a data distribution similar to that of fingpt-sentiment, closely mirrors the results obtained on the latter. For the wikitext dataset, the rendition of Embed Parrot tailored for Llama2 remains comparably stable, despite a modest decline in certain metrics. In the case of ChatGLM-6B, although there is a notable decrease in the Rouge metric scores, the F1 and Sentence-Bert scores continue to be relatively high. This can be attributed to the model's ability to recover tokens that, while not being exact replicas of the original, still bear a high degree of semantic resemblance.
In an effort to induce the language models to generate more coherent outputs, we have incorporated the Perplexity (PPL) provided by the model into the loss function. However, in most instances, the addition of PPL did not perform the anticipated optimization function. We hypothesize that this is due to the relatively short average token length of the datasets employed, rendering PPL statistically insignificant. Nevertheless, in the case of the wikitext dataset with token sequences exceeding 512, we found that integrating PPL can enhance semantic similarity to a certain degree.

\subsubsection{Impacts of Sequence Length}
We systematically analyzed the influence of token length on Embed Parrot across three distinct datasets, with the results presented in Table \ref{length-fingpt-sentiment}. We observe that for ChatGLM-6B, Embed Parrot exhibits poorer performance in reconstructing shorter texts when compared to longer ones. We hypothesize that this is due to ChatGLM-6B's utilization of the GLM architecture, which captures the hidden relationships between tokens more effectively. Consequently, embeddings of longer token sequences contain richer semantic information which facilitates reconstruction by Embed Parrot. In the case of Llama2-7B, the impact of token length is relatively minimal, with performance remaining consistent across various token length intervals. Although there is a noticeable decline in ROUGE and F1 scores within the token length range of [256,512), the Sentence-BERT semantic scores do not exhibit a significant decrease. This suggests that while the tokens reconstructed by Embed Parrot may not always be the most accurate, they tend to be semantically similar (e.g., "People" being reconstructed as 'Person').
Further analysis of our results on other datasets has been conducted and can be found in the Appendix \ref{append:length}.

\subsection{Denfense}
\begin{table*}
\centering
\caption{
 Effectiveness of defense strategy.
"Layer" indicates the layer at which the defensive transformation is applied.
}
\label{table:defense-result}
\begin{tabular}{llccccccc}
\toprule
Model &  Layer& ROUGE-1 & ROUGE-L & F1 Score & Sentence-BERT  & PPL(origin/defense) \\
\midrule
\multirow[c]{2}{*}{ChatGLM-6B} & 0 & 61.58 & 61.54 & 43.23 & 42.82 & 6.17/9.00 \\
 & 28 & 26.98 & 26.08 & 18.78 & 29.52 & 6.17/10.26 \\
\midrule
\multirow[c]{2}{*}{Llama2-7B} & 0 & 8.96 & 8.93 & 3.21 & 11.62 & 5.44/8.85\\
 & 30 & 16.48 & 16.00 & 10.19 & 23.51 & 5.44/9.13 \\
\bottomrule
\end{tabular}
\end{table*}

\subsubsection{Denfense Strategy}\label{defense-strategy}
Inspired by \cite{Mi2023PrivacyPreservingFR} and \cite{Wang_Liu_Luo_Yang_Wang_2022}, we introduce a novel data transformation technique that aims to protect the privacy of input embeddings. The method utilizes an overlap matrix in conjunction with the Discrete Cosine Transform (DCT) and its inverse (IDCT) to obfuscate the sensitive features of the embeddings.

Given an input embedding $\mathbf{E} \in \mathbb{R}^{n \times d}$, where $n$ is the number of embeddings and $d$ is the dimensionality.
We begin by generating a random permutation of indices from the set $\{1, 2, \ldots, d\}$
and partitioning it into $K$ subsets. Each subset corresponds to a unique overlap matrix $\mathbf{O}_i \in \{0, 1\}^{d \times d}$, constructed as follows:
(1) Identify the non-zero indices in each subset, denoted by a subset vector $\mathbf{v}_i$.
(2) Initialize a zero matrix $\mathbf{M}_i = \mathbf{0}_{d \times d}$.
(3) For each non-zero index $j$ in the subset vector $\mathbf{v}_i$, excluding the last index, set $\mathbf{M}_{i}[v_{i_j}, v_{i_j}] = 1$ and $\mathbf{M}_{i}[v_{i_j}, v_{i_{j+1}}] = 1$ to embed the overlap pattern.
(4) Transpose $\mathbf{M}_i$ to obtain the final overlap matrix, $\mathbf{O}_i = \mathbf{M}_i^T$.
Subsequently, we randomly select one of the overlap matrices $\mathbf{O}_s$ and apply the following transformation to the input embeddings by
$\mathbf{E}_{\operatorname{trans}} = \operatorname{IDCT}(\operatorname{DCT}(\mathbf{E}^T) \cdot \mathbf{O}_s)^T$
This transformation leverages the frequency domain to disperse the original embedding information across the spectrum, and then selectively retains and drops information according to the binary pattern in $\mathbf{O}_s$. As a result, $\mathbf{E}_{\operatorname{trans}}$ contains input embeddings with altered characteristics, which impedes the recovery of sensitive data while allowing for effective model forwarding.
The overlap matrices are saved for repeatability in the transformation process. This ensures consistent application of the method across different phases of model development and evaluation.

\subsubsection{Defense Resutls}
As illustrated in Table \ref{table:defense-result}, for ChatGLM-6B, applying the transformation delineated in \ref{defense-strategy} to the hidden states output at the 28th layer is more effective at diminishing the reconstructive capability of Embed Parrot than applying the same transformation to the hidden states output at the initial layer (layer 0). However, this approach also impacts the inherent generative performance of the model to a certain extent, resulting in an increase in ChatGLM-6B's perplexity (PPL) from 6.17 to 10.26. In contrast, for Llama2-7B, transforming the initial layer's output not only reduces the efficacy of Embed Parrot in terms of reconstruction but also inflicts marginally less damage on the model's origin generative performance. The differences between ChatGLM-6B and Llama2-7B are once again reflected, even in the context of their defensive measures.




\section{Conclusions and Future Directions}
\label{sec:con}
Our proposed Embed Parrot has proven to be a viable method for restoring the hidden states of a specified layer to the original input on both ChatGLM-6B and Llama2-7B, without exhibiting significant influence from token length or being confined to the training dataset utilized. While we have endeavored to conduct a comprehensive set of experiments to substantiate the practicality and feasibility of our approach, some limitations persist due to constraints in time and resources. For instance, we identified a particularly challenging layer for reconstruction using other methods in LLMs and posited that if Embed Parrot could achieve commendable metrics on this difficult layer, it would presumably perform even better on other layers; this hypothesis, however, remains unverified. In future work, we will validate the performance of Embed Parrot using shallower hidden states. Additionally, the network architecture employed for Embed Parrot is relatively simple, which may raise concerns regarding its generalizability. We plan to explore optimal structures for Embed Parrot by introducing more parameters, incorporating regularization techniques, and other means. Moreover, we aim to extend our research to training on a broader array of datasets for additional epochs to delve deeper into the learning capacity and generalization performance of Embed Parrot.

\bibliographystyle{icml2024}
\bibliography{ref}

\begin{thebibliography}{35}
\providecommand{\natexlab}[1]{#1}
\providecommand{\url}[1]{\texttt{#1}}
\expandafter\ifx\csname urlstyle\endcsname\relax
  \providecommand{\doi}[1]{doi: #1}\else
  \providecommand{\doi}{doi: \begingroup \urlstyle{rm}\Url}\fi

\bibitem[Babakniya et~al.(2023)Babakniya, Elkordy, Ezzeldin, Liu, Song, El-Khamy, and Avestimehr]{babakniya2023slora}
Babakniya, S., Elkordy, A.~R., Ezzeldin, Y.~H., Liu, Q., Song, K.-B., El-Khamy, M., and Avestimehr, S.
\newblock Slora: Federated parameter efficient fine-tuning of language models, 2023.

\bibitem[Balunovic et~al.(2022)Balunovic, Dimitrov, Jovanovi{\'c}, and Vechev]{balunovic2022lamp}
Balunovic, M., Dimitrov, D.~I., Jovanovi{\'c}, N., and Vechev, M.
\newblock {LAMP}: Extracting text from gradients with language model priors.
\newblock In Oh, A.~H., Agarwal, A., Belgrave, D., and Cho, K. (eds.), \emph{Advances in Neural Information Processing Systems}, 2022.

\bibitem[Borgeaud et~al.(2022)Borgeaud, Mensch, Hoffmann, Cai, Rutherford, Millican, van~den Driessche, Lespiau, Damoc, Clark, de~Las~Casas, Guy, Menick, Ring, Hennigan, Huang, Maggiore, Jones, Cassirer, Brock, Paganini, Irving, Vinyals, Osindero, Simonyan, Rae, Elsen, and Sifre]{borgeaud2022improving}
Borgeaud, S., Mensch, A., Hoffmann, J., Cai, T., Rutherford, E., Millican, K., van~den Driessche, G., Lespiau, J.-B., Damoc, B., Clark, A., de~Las~Casas, D., Guy, A., Menick, J., Ring, R., Hennigan, T., Huang, S., Maggiore, L., Jones, C., Cassirer, A., Brock, A., Paganini, M., Irving, G., Vinyals, O., Osindero, S., Simonyan, K., Rae, J.~W., Elsen, E., and Sifre, L.
\newblock Improving language models by retrieving from trillions of tokens, 2022.

\bibitem[Chen et~al.(2023)Chen, Feng, Zhou, Yin, and Zheng]{chen2023federated}
Chen, C., Feng, X., Zhou, J., Yin, J., and Zheng, X.
\newblock Federated large language model: A position paper, 2023.

\bibitem[Du et~al.(2022)Du, Qian, Liu, Ding, Qiu, Yang, and Tang]{du2022glm}
Du, Z., Qian, Y., Liu, X., Ding, M., Qiu, J., Yang, Z., and Tang, J.
\newblock Glm: General language model pretraining with autoregressive blank infilling.
\newblock In \emph{Proceedings of the 60th Annual Meeting of the Association for Computational Linguistics (Volume 1: Long Papers)}, pp.\  320--335, 2022.

\bibitem[Fowl et~al.(2023)Fowl, Geiping, Reich, Wen, Czaja, Goldblum, and Goldstein]{fowl2023decepticons}
Fowl, L.~H., Geiping, J., Reich, S., Wen, Y., Czaja, W., Goldblum, M., and Goldstein, T.
\newblock Decepticons: Corrupted transformers breach privacy in federated learning for language models.
\newblock In \emph{The Eleventh International Conference on Learning Representations}, 2023.

\bibitem[Geva et~al.(2021)Geva, Schuster, Berant, and Levy]{geva2021transformer}
Geva, M., Schuster, R., Berant, J., and Levy, O.
\newblock Transformer feed-forward layers are key-value memories, 2021.

\bibitem[Gu et~al.(2023)Gu, Kabir, Ramsurrun, Vosoughi, and Mehnaz]{Gu2023TowardsSL}
Gu, K., Kabir, E., Ramsurrun, N., Vosoughi, S., and Mehnaz, S.
\newblock Towards sentence level inference attack against pre-trained language models.
\newblock \emph{Proc. Priv. Enhancing Technol.}, 2023:\penalty0 62--78, 2023.

\bibitem[Kiros et~al.(2015)Kiros, Zhu, Salakhutdinov, Zemel, Torralba, Urtasun, and Fidler]{kiros2015skipthought}
Kiros, R., Zhu, Y., Salakhutdinov, R., Zemel, R.~S., Torralba, A., Urtasun, R., and Fidler, S.
\newblock Skip-thought vectors, 2015.

\bibitem[Le \& Mikolov(2014)Le and Mikolov]{le2014distributed}
Le, Q.~V. and Mikolov, T.
\newblock Distributed representations of sentences and documents, 2014.

\bibitem[Li et~al.(2023)Li, Xu, and Song]{li-etal-2023-sentence}
Li, H., Xu, M., and Song, Y.
\newblock Sentence embedding leaks more information than you expect: Generative embedding inversion attack to recover the whole sentence.
\newblock In Rogers, A., Boyd-Graber, J., and Okazaki, N. (eds.), \emph{Findings of the Association for Computational Linguistics: ACL 2023}, pp.\  14022--14040, Toronto, Canada, July 2023. Association for Computational Linguistics.
\newblock \doi{10.18653/v1/2023.findings-acl.881}.

\bibitem[Lin(2004)]{lin-2004-rouge}
Lin, C.-Y.
\newblock {ROUGE}: A package for automatic evaluation of summaries.
\newblock In \emph{Text Summarization Branches Out}, pp.\  74--81, Barcelona, Spain, July 2004. Association for Computational Linguistics.

\bibitem[Merity et~al.(2016)Merity, Xiong, Bradbury, and Socher]{merity2016pointer}
Merity, S., Xiong, C., Bradbury, J., and Socher, R.
\newblock Pointer sentinel mixture models, 2016.

\bibitem[Mi et~al.(2023)Mi, Huang, Ji, Zhao, Wu, Xu, Ding, and Zhou]{Mi2023PrivacyPreservingFR}
Mi, Y., Huang, Y., Ji, J.-B., Zhao, M., Wu, J., Xu, X., Ding, S., and Zhou, S.
\newblock Privacy-preserving face recognition using random frequency components.
\newblock \emph{2023 IEEE/CVF International Conference on Computer Vision (ICCV)}, pp.\  19616--19627, 2023.

\bibitem[Morris et~al.(2023{\natexlab{a}})Morris, Kuleshov, Shmatikov, and Rush]{morris2023text}
Morris, J.~X., Kuleshov, V., Shmatikov, V., and Rush, A.~M.
\newblock Text embeddings reveal (almost) as much as text, 2023{\natexlab{a}}.

\bibitem[Morris et~al.(2023{\natexlab{b}})Morris, Zhao, Chiu, Shmatikov, and Rush]{morris2023language}
Morris, J.~X., Zhao, W., Chiu, J.~T., Shmatikov, V., and Rush, A.~M.
\newblock Language model inversion, 2023{\natexlab{b}}.

\bibitem[Naveed et~al.(2023)Naveed, Khan, Qiu, Saqib, Anwar, Usman, Akhtar, Barnes, and Mian]{naveed2023comprehensive}
Naveed, H., Khan, A.~U., Qiu, S., Saqib, M., Anwar, S., Usman, M., Akhtar, N., Barnes, N., and Mian, A.
\newblock A comprehensive overview of large language models, 2023.

\bibitem[Pan et~al.(2020)Pan, Zhang, Ji, and Yang]{9152761}
Pan, X., Zhang, M., Ji, S., and Yang, M.
\newblock Privacy risks of general-purpose language models.
\newblock In \emph{2020 IEEE Symposium on Security and Privacy (SP)}, pp.\  1314--1331, 2020.
\newblock \doi{10.1109/SP40000.2020.00095}.

\bibitem[Phong et~al.(2017)Phong, Aono, Hayashi, Wang, and Moriai]{phong2017privacy}
Phong, L.~T., Aono, Y., Hayashi, T., Wang, L., and Moriai, S.
\newblock Privacy-preserving deep learning: Revisited and enhanced.
\newblock In \emph{Applications and Techniques in Information Security: 8th International Conference, ATIS 2017, Auckland, New Zealand, July 6--7, 2017, Proceedings}, pp.\  100--110. Springer, 2017.

\bibitem[Radford et~al.(2019)Radford, Wu, Child, Luan, Amodei, and Sutskever]{radford2019language}
Radford, A., Wu, J., Child, R., Luan, D., Amodei, D., and Sutskever, I.
\newblock Language models are unsupervised multitask learners.
\newblock 2019.

\bibitem[Rajpurkar et~al.(2016)Rajpurkar, Zhang, Lopyrev, and Liang]{rajpurkar2016squad}
Rajpurkar, P., Zhang, J., Lopyrev, K., and Liang, P.
\newblock Squad: 100,000+ questions for machine comprehension of text, 2016.

\bibitem[Reimers \& Gurevych(2019)Reimers and Gurevych]{reimers2019sentencebert}
Reimers, N. and Gurevych, I.
\newblock Sentence-bert: Sentence embeddings using siamese bert-networks, 2019.

\bibitem[Shokri et~al.(2017)Shokri, Stronati, Song, and Shmatikov]{shokri2017membership}
Shokri, R., Stronati, M., Song, C., and Shmatikov, V.
\newblock Membership inference attacks against machine learning models, 2017.

\bibitem[Song \& Raghunathan(2020)Song and Raghunathan]{song2020information}
Song, C. and Raghunathan, A.
\newblock Information leakage in embedding models, 2020.

\bibitem[Touvron et~al.(2023)Touvron, Martin, Stone, Albert, Almahairi, Babaei, Bashlykov, Batra, Bhargava, Bhosale, Bikel, Blecher, Ferrer, Chen, Cucurull, Esiobu, Fernandes, Fu, Fu, Fuller, Gao, Goswami, Goyal, Hartshorn, Hosseini, Hou, Inan, Kardas, Kerkez, Khabsa, Kloumann, Korenev, Koura, Lachaux, Lavril, Lee, Liskovich, Lu, Mao, Martinet, Mihaylov, Mishra, Molybog, Nie, Poulton, Reizenstein, Rungta, Saladi, Schelten, Silva, Smith, Subramanian, Tan, Tang, Taylor, Williams, Kuan, Xu, Yan, Zarov, Zhang, Fan, Kambadur, Narang, Rodriguez, Stojnic, Edunov, and Scialom]{touvron2023llama}
Touvron, H., Martin, L., Stone, K., Albert, P., Almahairi, A., Babaei, Y., Bashlykov, N., Batra, S., Bhargava, P., Bhosale, S., Bikel, D., Blecher, L., Ferrer, C.~C., Chen, M., Cucurull, G., Esiobu, D., Fernandes, J., Fu, J., Fu, W., Fuller, B., Gao, C., Goswami, V., Goyal, N., Hartshorn, A., Hosseini, S., Hou, R., Inan, H., Kardas, M., Kerkez, V., Khabsa, M., Kloumann, I., Korenev, A., Koura, P.~S., Lachaux, M.-A., Lavril, T., Lee, J., Liskovich, D., Lu, Y., Mao, Y., Martinet, X., Mihaylov, T., Mishra, P., Molybog, I., Nie, Y., Poulton, A., Reizenstein, J., Rungta, R., Saladi, K., Schelten, A., Silva, R., Smith, E.~M., Subramanian, R., Tan, X.~E., Tang, B., Taylor, R., Williams, A., Kuan, J.~X., Xu, P., Yan, Z., Zarov, I., Zhang, Y., Fan, A., Kambadur, M., Narang, S., Rodriguez, A., Stojnic, R., Edunov, S., and Scialom, T.
\newblock Llama 2: Open foundation and fine-tuned chat models, 2023.

\bibitem[van Aken et~al.(2020)van Aken, Winter, Löser, and Gers]{vanaken2020visbert}
van Aken, B., Winter, B., Löser, A., and Gers, F.~A.
\newblock Visbert: Hidden-state visualizations for transformers, 2020.

\bibitem[Vaswani et~al.(2017)Vaswani, Shazeer, Parmar, Uszkoreit, Jones, Gomez, Kaiser, and Polosukhin]{vaswani2017attention}
Vaswani, A., Shazeer, N., Parmar, N., Uszkoreit, J., Jones, L., Gomez, A.~N., Kaiser, {\L}., and Polosukhin, I.
\newblock Attention is all you need.
\newblock \emph{Advances in neural information processing systems}, 30, 2017.

\bibitem[Wang et~al.(2022)Wang, Liu, Luo, Yang, and Wang]{Wang_Liu_Luo_Yang_Wang_2022}
Wang, Y., Liu, J., Luo, M., Yang, L., and Wang, L.
\newblock Privacy-preserving face recognition in the frequency domain.
\newblock \emph{Proceedings of the AAAI Conference on Artificial Intelligence}, 36\penalty0 (3):\penalty0 2558--2566, Jun. 2022.
\newblock \doi{10.1609/aaai.v36i3.20157}.

\bibitem[Yang et~al.(2023)Yang, Liu, and Wang]{yang2023fingpt}
Yang, H., Liu, X.-Y., and Wang, C.~D.
\newblock Fingpt: Open-source financial large language models, 2023.

\bibitem[Yao et~al.(2023)Yao, Zhao, Yu, Du, Shafran, Narasimhan, and Cao]{yao2023react}
Yao, S., Zhao, J., Yu, D., Du, N., Shafran, I., Narasimhan, K., and Cao, Y.
\newblock React: Synergizing reasoning and acting in language models, 2023.

\bibitem[Zhang et~al.(2023{\natexlab{a}})Zhang, Vahidian, Kuo, Li, Zhang, Wang, and Chen]{zhang2023building}
Zhang, J., Vahidian, S., Kuo, M., Li, C., Zhang, R., Wang, G., and Chen, Y.
\newblock Towards building the federated gpt: Federated instruction tuning, 2023{\natexlab{a}}.

\bibitem[Zhang et~al.(2023{\natexlab{b}})Zhang, Yang, Dai, Qu, and Xu]{zhang2023federated}
Zhang, Z., Yang, Y., Dai, Y., Qu, L., and Xu, Z.
\newblock When federated learning meets pre-trained language models' parameter-efficient tuning methods, 2023{\natexlab{b}}.

\bibitem[Zhao et~al.(2020)Zhao, Mopuri, and Bilen]{zhao2020idlg}
Zhao, B., Mopuri, K.~R., and Bilen, H.
\newblock idlg: Improved deep leakage from gradients, 2020.

\bibitem[Zhu et~al.(2019{\natexlab{a}})Zhu, Liu, and Han]{NEURIPS2019_60a6c400}
Zhu, L., Liu, Z., and Han, S.
\newblock Deep leakage from gradients.
\newblock In Wallach, H., Larochelle, H., Beygelzimer, A., d\textquotesingle Alch\'{e}-Buc, F., Fox, E., and Garnett, R. (eds.), \emph{Advances in Neural Information Processing Systems}, volume~32. Curran Associates, Inc., 2019{\natexlab{a}}.

\bibitem[Zhu et~al.(2019{\natexlab{b}})Zhu, Liu, and Han]{zhu2019deep}
Zhu, L., Liu, Z., and Han, S.
\newblock Deep leakage from gradients, 2019{\natexlab{b}}.

\end{thebibliography}

\clearpage

\section{Appendix}
\label{sec:apend}

\subsection{Impacts of Fine-tuning} \label{impacts-of-FT}
Membership inference attacks\cite{shokri2017membership} exploit a certain insight that machine learning models often behave differently on the data on which they are trained than on which they were first seen. This raises the question of whether data is easier to recover with a model trained on it. Therefore, we used LoRA to fine-tune the model on the specific datasets and ultimately evaluate the impact of fine-tuning on restoring the embeddings to their original inputs. 

For ChatGLM-6B, we fine-tune its upscale/downscale projection layers in Multilayer Perceptron(MLP)
with LoRA on the dataset Fingpt/fingpt-sentiment for four epochs, Finally, the fine-tuned model is used to test the reduction performance on the training data set it used. For Llama2-7B, we employ similar fine-tuning settings as those utilized for ChatGLM-6B. However, the specific layers adjusted with the LoRA fine-tuning technique for Llama2-7B are the query projection layers, key projection layers, value projection layers, and output projection layers. These layers correspond to the components of the attention mechanism that transform the input representations into the respective query, key, and value spaces for the attention calculations, as well as the subsequent mapping to the output space. 

According to the experimental results shown in Table \ref{BEI:ChatGLM-and-LLama} and Table \ref{HEI:ChatGLM-and-LLama}, we can find that using LoRA fine-tuning on ChatGLM-6B does help to improve the Rouge score, but the improvement effect is not obvious on most layers, only in the last few layers.
we can also find that using  LoRA fine-tuning on Llama-7b-chat-hf hurts almost all layers and makes it score lower than the baseline, especially in the last few layers.
The performance of ChatGLM markedly diverges from that of Llama2.
As a result, we believe that fine-tuning is of limited use and also depends to some extent on the architecture of the pre-trained model.

\subsection{Impacts Of Sequence Length}\label{append:length}
As shown in Table \ref{length-wikitext},a point of interest is the remarkable reconstruction performance of Embed Parrot on Llama2-7b for tokens in the (0,8) length range on the Wikitext dataset, despite not being trained on it. We speculate that this is attributable to Llama2-7b having 'memorized' content from Wikitext during the pre-training phase. When token lengths are short, semantic information is subject to less 'interference', thus aiding reconstruction by Embed Parrot.
Furthermore, we discovered that although incorporating PPL into the loss function generally exerts a slight detrimental effect on Embed Parrot in the majority of cases, it enhances the Sentence-BERT scores for longer token intervals (512,1024). Hence, it is plausible to consider PPL as a potentially beneficial loss metric for Embed Parrot when dealing with extended token sequences.

\begin{table*}
\centering
\caption{
Performance of Embed Parrot to reconstruct tokens with different length ranges on the fingpt-fiqa\_qa dataset. 
}
\begin{tabular}{llccccccccccc}
\toprule
 Model &  & Metrics & (0, 8) & [8, 16) & [16, 32) & [32, 64)  \\
\midrule
\multirow[c]{4}{*}{ChatGLM-6B} &\multirow[c]{4}{*} & ROUGE-1 & 63.71 & 73.26 & 77.51 & 81.50  \\
 & &ROUGE-L & 63.71 & 73.22 & 77.43 & 81.50  \\
 & &F1 Score & 57.25 & 68.75 & 71.24 & 75.17 \\
 & &Sentence-BERT & 67.00 & 75.31 & 77.45 & 80.12 \\
\midrule
\multirow[c]{4}{*}{Llama2-7B(CP)}&\multirow[c]{4}{*} & ROUGE-1 & 65.45 & 65.43 & 64.93 & 64.71\\
 & &ROUGE-L & 65.45 & 65.40 & 64.90 & 64.42\\
 & &F1 Score & 64.24 & 62.77 & 60.56 & 59.22 \\
 & &Sentence-BERT  & 70.38 & 67.73 & 68.49 & 68.84 \\
\midrule
\multirow[c]{4}{*}{Llama2-7B(LP)}&\multirow[c]{4}{*} & ROUGE-1 & 66.11 & 65.29 & 64.67 & 64.63 \\
 & &ROUGE-L & 66.11 & 65.28 & 64.64 & 64.34 \\
 & &F1 Score & 65.30 & 62.61 & 60.26 & 59.14 \\
 & &Sentence-BERT  & 70.85 & 67.42 & 67.70 & 68.15 \\
\midrule
\multirow[c]{4}{*}{Llama2-7B}&\multirow[c]{4}{*} & ROUGE-1 & 66.78 & 66.04 & 65.92 & 65.22 \\
 & &ROUGE-L & 66.78 & 66.02 & 65.89 & 65.22 \\
 & &F1 Score & 65.71 & 63.37 & 61.63 & 59.61 \\
 & &Sentence-BERT  & 71.29 & 68.20 & 69.19 & 69.48 \\
\bottomrule
\end{tabular}\label{length-fingpt-fiqa-qa}
\end{table*}

\begin{table*}
\centering
\caption{Performance of EP to reconstruct tokens with different length ranges on Wikitext dataset. Owing to tokenizer differences, ChatGLM-6B did not process data beyond 512 tokens, while Llama2-7B processed a few instances.
}

\begin{tabular}{lllcccccccccc}
\toprule
 Model & & Metrics &(0, 8) & [8, 16) & [16, 32) & [32, 64) & [64, 128) & [128, 256) & [256, 512) & [512, 1024)  
 \\
\midrule
\multirow[c]{4}{*}{ChatGLM-6B}&\multirow[c]{4}{*} & ROUGE-1 & 39.41 & 46.53 & 57.94 & 66.38 & 65.81 & 67.29 & 68.06 & - 
\\
 & &ROUGE-L & 39.41 & 46.53 & 57.92 & 66.27 & 65.63 & 67.07 & 67.72 & - 
 \\
 & &F1 Score & 25.86 & 38.64 & 48.86 & 54.28 & 55.07 & 57.12 & 58.13 & - 
 \\
 & &Sentence-BERT  & 38.30 & 43.36 & 50.93 & 49.61 & 47.36 & 49.62 & 53.14 & - 
 \\
\midrule
\multirow[c]{4}{*}{Llama2-7B(CP)}&\multirow[c]{4}{*} & ROUGE-1 & 98.60 & 47.36 & 49.26 & 49.61 & 47.60 & 48.29 & 47.91 & 48.55 
\\
 & &ROUGE-L & 98.60 & 47.36 & 49.26 & 49.36 & 47.26 & 47.75 & 47.04 & 47.14 
 \\
 & &F1 Score & 98.60 & 46.79 & 46.91 & 45.14 & 42.46 & 42.82 & 42.41 & 43.28 
 \\
 & &Sentence-BERT  & 98.90 & 52.19 & 48.19 & 40.21 & 33.44 & 33.37 & 33.90 & 42.06 
 \\
\midrule
\multirow[c]{4}{*}{Llama2-7B(LP)}&\multirow[c]{4}{*} & ROUGE-1 & 98.55 & 46.73 & 49.35 & 49.85 & 47.48 & 48.36 & 47.81 & 48.58 
\\
 & &ROUGE-L & 98.55 & 46.73 & 49.35 & 49.75 & 47.13 & 47.84 & 46.97 & 47.10 
 \\
 & &F1 Score & 98.55 & 46.13 & 47.05 & 45.34 & 42.28 & 42.83 & 42.25 & 43.29 
 \\
 & &Sentence-BERT  & 98.86 & 51.64 & 48.28 & 40.74 & 33.28 & 33.19 & 33.23 & 38.91 
 \\
\midrule
\multirow[c]{4}{*}{Llama2-7B}&\multirow[c]{4}{*} & ROUGE-1 & 98.58 & 47.69 & 49.78 & 50.66 & 48.29 & 49.35 & 48.79 & 48.89 
\\
 & &ROUGE-L & 98.58 & 47.69 & 49.71 & 50.61 & 48.00 & 48.86 & 48.04 & 47.80 
 \\
 & &F1 Score & 98.58 & 47.11 & 47.40 & 46.26 & 43.16 & 43.91 & 43.38 & 43.48 
 \\
 & &Sentence-BERT  & 98.87 & 52.53 & 48.19 & 40.90 & 33.84 & 33.14 & 33.23 & 37.57 
 \\
\bottomrule
\end{tabular}\label{length-wikitext}
\end{table*}

\begin{table*}
\centering
\caption{The performance of the Base Embed Inversion (BEI) attack method on models both before and after fine-tuning(FT). 
The numbers in the first row indicate the layer indexes. 
}

\label{BEI:ChatGLM-and-LLama}
\begin{tabular}{llcccccccc}
\toprule
Model & Metrics  & 0 & 5 & 10 & 15 & 20 & 25 & last \\
\midrule
\multirow[c]{4}{*}{ChatGLM-6B} & F1 Score & 70.75 & 68.56 & 67.31 & 74.10 & 65.38 & 49.49 & 39.19 \\
 & ROUGE-1 & 100.00 & 97.41 & 93.26 & 86.62 & 72.33 & 55.46 & 42.86 \\
 & ROUGE-L & 100.00 & 97.47 & 93.29 & 86.64 & 72.41 & 53.16 & 39.42 \\
 & Sentence-BERT  & 71.64 & 70.61 & 70.48 & 77.92 & 71.51 & 64.41 & 53.46 \\
\midrule
\multirow[c]{4}{*}{FT ChatGLM-6B} & F1 Score & 70.75 & 69.73 & 67.64 & 71.85 & 63.46 & 48.01 & 51.21 \\
 & ROUGE-1 & 100.00 & 98.94 & 93.54 & 87.31 & 72.69 & 57.46 & 54.81 \\
 & ROUGE-L & 100.00 & 98.95 & 93.56 & 87.31 & 72.72 & 55.08 & 52.48 \\
 & Sentence-BERT  & 71.64 & 71.13 & 70.79 & 77.53 & 68.64 & 64.19 & 66.99 \\
\midrule
\multirow[c]{4}{*}{Llama2-7B} & F1 Score & 0.17 & 1.98 & 2.46 & 3.25 & 13.10 & 18.10  & 23.01 \\
 & ROUGE-1 & 0.33 & 1.71 & 1.75 & 3.20 & 9.33 & 13.34  & 18.23 \\
 & ROUGE-L & 0.33 & 1.68 & 1.69 & 3.04 & 8.57 & 12.19  & 17.59 \\
 & Sentence-BERT & 3.00 & 2.31 & 2.99 & 6.62 & 22.61 & 30.95  & 32.41 \\
\midrule
\multirow[c]{4}{*}{FT Llama2-7B} & F1 Score & 0.17 & 1.61 & 1.43 & 2.98 & 12.45 & 18.35  & 4.84 \\
 & ROUGE-1 & 0.33 & 1.81 & 1.24 & 2.49 & 8.75 & 13.35  & 7.35 \\
 & ROUGE-L & 0.33 & 1.79 & 1.20 & 2.39 & 7.95 & 12.06  & 7.10 \\
 & Sentence-BERT & 3.00 & 1.91 & 2.34 & 8.16 & 22.85 & 28.02 & 7.23 \\
\bottomrule
\end{tabular}
\end{table*}

\begin{table*}
\centering
\caption{The performance of the Hotmap Embed Inversion (HEI) attack method on models both before and after fine-tuning(FT)
The numbers in the first row indicate the layer indexes. 
}
\label{HEI:ChatGLM-and-LLama}
\begin{tabular}{llcccccccc}
\toprule
Model & Metrics  & 0 & 5 & 10 & 15 & 20 & 25 & last \\
\midrule
\multirow[c]{4}{*}{ChatGLM-6B} & F1 Score & 99.94 & 96.46 & 93.94 & 91.23 & 60.68 & 51.78 & 40.23 \\
 & ROUGE-1 & 100.00 & 96.72 & 92.17 & 89.89 & 61.05 & 55.28 & 41.37 \\
 & ROUGE-L & 100.00 & 96.76 & 92.18 & 89.87 & 60.98 & 53.69 & 38.72 \\
 & Sentence-BERT  & 99.99 & 98.15 & 97.01 & 95.07 & 67.03 & 71.28 & 64.29 \\
\midrule
\multirow[c]{4}{*}{FT ChatGLM-6B} & F1 Score & 99.94 & 96.41 & 94.36 & 90.28 & 67.30 & 52.72 & 50.34 \\
 & ROUGE-1 & 100.00 & 96.68 & 92.91 & 90.15 & 68.34 & 56.32 & 54.48 \\
 & ROUGE-L & 100.00 & 96.72 & 92.89 & 90.10 & 68.27 & 54.61 & 52.41 \\
 & Sentence-BERT  & 99.99 & 98.14 & 97.32 & 93.19 & 69.79 & 67.84 & 71.91 \\
\midrule
\multirow[c]{4}{*}{Llama2-7B} & F1 Score & 85.07 & 73.44 & 65.39 & 61.73 & 53.50 & 39.30 &  16.67 \\
 & ROUGE-1 & 65.22 & 58.93 & 53.09 & 50.11 & 44.38 & 34.39 &  19.51 \\
 & ROUGE-L & 65.23 & 58.95 & 53.07 & 50.04 & 44.40 & 34.37 &  19.13 \\
 & Sentence-BERT  & 90.15 & 84.19 & 80.27 & 77.75 & 72.65 & 61.01 &  25.11 \\
\midrule
\multirow[c]{4}{*}{FT Llama2-7B} & F1 Score & 85.07 & 73.43 & 64.02 & 59.99 & 53.78 & 39.53 &  4.67 \\
 & ROUGE-1 & 65.22 & 58.82 & 52.34 & 49.45 & 44.35 & 33.45 &  6.17 \\
 & ROUGE-L & 65.23 & 58.84 & 52.28 & 49.41 & 44.30 & 33.40 &  6.13 \\
 & Sentence-BERT  & 90.15 & 84.20 & 79.25 & 76.35 & 72.56 & 60.01 &  8.95 \\
\bottomrule
\end{tabular}
\end{table*}

\end{document}